\RequirePackage{fix-cm}
\documentclass[smallextended]{svjour3}       
\smartqed  
\usepackage{graphicx}

\usepackage{graphicx}
\usepackage{amsmath}
\usepackage{amssymb}

\usepackage{algorithmic}
\usepackage{algorithm}
\usepackage{subfigure}
\usepackage{array}
\usepackage{multirow,tabularx}
\usepackage{lineno}
\usepackage{pdfsync}
\usepackage{multicol}
\usepackage{color}

\newcommand{\bfz}{{\textbf{z}}}
\newcommand{\bfx}{{\textbf{x}}}

\newcommand{\bfomega}{{\boldsymbol{\omega}}}

\begin{document}

\title{Supervised multiview learning based on simultaneous learning of multiview intact and single view classifier}

\author{Qingjun Wang\and
Haiyan Lv\and
Jun Yue\and
Eugene Mitchell}

\institute{Qingjun Wang, Jun Yue\at
School of Information and Electrical Engineering, Ludong University, Yantai 264025, China\\
\email{qjwang386@hotmail.com}
\and
Haiyan Lv\at
Naval Aeronautical and Astronautical University, Yantai 264025,China
\and
Eugene Mitchell\at
Department of Computer Science, Ryerson University, Toronto, ON M5B 2K3, Canada\\
\email{emitchell328496@outlook.com}
}

\date{Received: date / Accepted: date}

\maketitle

\begin{abstract}
Multiview learning problem refers to the problem of learning a classifier from multiple view data. In this data set, each data points is presented by multiple different views. In this paper, we propose a novel method for this problem. This method is based on two assumptions. The first assumption is that each data point has an intact feature vector, and each view is obtained by a linear transformation from the intact vector. The second assumption is that the intact vectors are discriminative, and in the intact space, we have a linear classifier to separate the positive class from the negative class. We define an intact vector for each data point, and a view-conditional transformation matrix for each view, and propose to reconstruct the multiple view feature vectors by the product of the corresponding intact vectors and transformation matrices. Moreover, we also propose a linear classifier in the intact space, and learn it jointly with the intact vectors. The learning problem is modeled by a minimization problem, and the objective function is composed of a Cauchy error estimator-based view-conditional reconstruction term over all data points and views, and a classification error term measured by hinge loss over all the intact vectors of all the data points. Some regularization terms are also imposed to different variables in the objective function. The minimization problem is solve by an iterative algorithm using alternate optimization strategy and gradient descent algorithm. The proposed algorithm shows it advantage in the compression to other multiview learning algorithms on benchmark data sets.
\end{abstract}

\section{Introduction}

\subsection{Background}

Multiview learning has been an important in machine learning community \cite{Wu2012661,Sun20132031,Yu20142431,Lu2015,Zha2015,FakeriTabrizi2015117,Liu20151233,liu2015supervised,sindhwani2008rkhs,wang2015regularized}. In traditional machine learning problems, we usually assume that a data point has a feature vector to represent its input information. For example, in image recognition problem, we can extract a visual feature vector from an image, using a texture descriptor \cite{Petrov20131499,Mohanty20131011,Mala201580,Yadav2015101,Luo2013709,Gui20143126,Wang2014,jiang2015manifold}. In this scene, the texture is a view of the image. However, there could be more than one view of an image. Besides the texture view, we can also extract feature vectors from other views, including shape and color. An other example is the problem of classification of scientific articles, and we may extract a feature vector from the main text of the article \cite{Long20151833,Hogenboom201546,Picard201595,La2015929,Hogenboom201546,Picard201595,Koopman2015,Chen2015473,Feng2015109,KumarNagwani20152589}. However, the main text is just one view the article, and we can also have features from other views, such as abstract, reference list, etc. Multiview learning argues that we should learn from more than one views to present the data and construct a predictor. The motive for multiview learning is that single view based data representation is usually incomplete, and different views can present complementary information for the learning problem. In the problem of multiview learning, the input of a data point is not just one single feature vector of one single view, but multiple feature vectors presenting different views. The target of multiview learning is to learn a predictor to take multiple view feature vectors to predict one single output of a data point. The problem of multiview learning can be classified to two types, supervised multiview learning and unsupervised learning.

\begin{itemize}
\item Supervised multiview learning refers to the problem of learning from a data set, where both the multiview input and output are available for each data point \cite{Li20142040,Jiang20141635,Hajmohammadi2014195}. In this problem, the output is usually a class label, or a continues response. In this case, the learning problem is to build a predictive model from the training data set to predict the output of a input data point, with help the input-output pairs of the training set.

\item Unsupervised multiview learning refers to the problem of cluster a set of data points, and the multiview inputs of each data point are given \cite{Feng2013343,Sublemontier2013,Zhao20137}. In this problem, the outputs of the data points are not available.
\end{itemize}

In this paper, we investigate the problem of supervised multiview learning, and propose a novel algorithm to solve it. The proposed method is based on an assumption that different views of a data point are generated from one single intact feature vector, and the view generation is performed by a linear transformation. We try to recover the intact feature vector for each data point from its multiview feature vectors, with guiding of its corresponding output, i.e., its binary class label.

\subsection{Relevant works}

There are some existing multiview learning methods. We the state-of-the-arts of them as follows.

\begin{itemize}
\item Zhang et al. \cite{Zhang2008752} proposed to use local learning (LL) method for the problem of multiview learning problem, and designs a local predictive model for each data point based on the multiview inputs. The local predictive model is learned on the nearest neighbors of a data point.

\item Sindhwani et al. \cite{sindhwani2005co} proposed to use co-training algorithm for multiview learning problems to improve the classification performance of each view (CT). This method is based on multiview regularization, and the agreement and smoothness over both labeled and unlabeled data points.

\item Quadrianto \cite{Quadrianto2011425} proposed a multiview learning algorithm to solve the problem of view disagreement (VD), i.e., different views of one single data point do not belong to the same class. This method uses a conditional entropy criterion to find the disagreement among different views, and remove the data points with view disagreement from the training set.

\item Zhai \cite{Zhai2012} proposed multiview metric learning method with global consistency and local smoothness (GL) for the multiview learning problem with partially labeled data set. This method simultaneously consider both the global consistency and local smoothness, by assuming that the different views has a shared latent feature space, and imposing global consistency and local structure to the learning procedure.

\item Chen et al. \cite{Chen20122365} proposed a statistical subspace multiview representation method (SS), by leveraging both multiview dependencies and supervision information. This method is based on a subspace Markov network of multiview latent, and assumes that the multiviews and the class labels are conditionally independent. The algorithm is based on the maximization of data likelihood, and the minimization of classification error.

\end{itemize}

\subsection{Contributions}

In this paper, we propose a novel supervised multiview learning method. This method is based on the assumption of single discriminative intact of different multiview inputs. Under this assumption, although there are different views of one single data point, one single intact feature vector exists for the data point. This intact feature vector is assumed to be discriminative, i.e., it can represents the class information of each data point. Moreover, the feature vector of each view of a data point can be obtained from the intact vector, by performing a linear view-conditional transformation to the intact feature vector. In this way, if we learn the discriminative intact feature vector for each training data point, we can learn a classifier in the intact with the help of the class labels of the training data points. To this end, we proposed a novel method to learn the hidden of the intact feature vector, the view-conditional transformation matrices, and the classifier in the intact space simultaneously. We define a intact feature vector for each data point, and a transformation matrix for each view. The feature vector of one view of each data point can be reconstructed as the product of its corresponding transformation matrix and intact feature vector. The reconstruction error for each view of each data point is measured by the Cauchy error estimator \cite{Idan20141108,Gallagher20151264}. To learn the optimal intact feature vectors and view-conditional transformation matrices, we propose to minimize the Cauchy errors. Moreover, due to the assumption that the intact feature vectors are discriminative, we also argue that we can design a classifier in the intact space, and the classifier can minimize the classification error. Thus we also propose to learn a linear classifier in the intact space, and use the hinge loss to measure the classification error the training set in the intact space \cite{Chen201580,Charuvaka201563}. To learn the optimal classifier parameter and the intact feature vectors, we also propose to minimize the hinge loss with regard to both the classifier parameter and the intact feature vectors.

To model the problem, we propose a joint optimization problem for learning of intact vectors, view-conditional transformation matrices, and the classifier parameter vector. The objective function of this problem is composed of two error terms, and three regularization terms. The firs error term is the view reconstruction term measured by Cauchy estimator over all the data points and views. The second error term is the classification error over all the intact feature vectors of all training data points, measured by hinge losses. The three regularization terms are all squared $\ell_2$ norm terms over each intact feature vectors, view-conditional matrices, and the classifier parameter vectors. The purpose of impose the squared $\ell_2$ norm to these variables are to reduce the complexity of the learned outputs. To minimize the proposed objective function, we adapt an alternate optimization strategy, i.e., when the objective function is minimized with regard to one variable, other variables are fixed. The optimization with regard to each variable is conducted by using gradient descent algorithm.

The contributions of this paper are of three parts:

\begin{enumerate}
\item We propose a novel supervised multiview learning framework by simultaneous learning of intact feature vectors, view-conditional transformation matrices, and classifier parameter vector.

\item We build a novel optimization problem for this learning problem, by considering both the view reconstruction problem, and the classifier learning problem.

\item We develop an iterative algorithm to solve this optimization problem based on alternate optimization strategy and gradient descent algorithm.
\end{enumerate}

\subsection{Paper organization}

This paper is organized as follows: In section \ref{sec:method}, the proposed method for supervised multiview learning is introduced. In this section, we first model this problem as a minimization problem of a objective function, and then solve it with an iterative algorithm. In section \ref{sec:experiments}, the proposed iterative algorithm is evaluated. We first give an analysis of its sensitivity to parameters, and then compare it to some state-of-the-art algorithms, and finally test the running time performance of the proposed algorithm. In section \ref{sec:conclusion}, we give the conclusion of this paper.

\section{Methods}
\label{sec:method}

In this section, we introduce the proposed supervised multiview learning method.

\subsection{Problem modeling}

We assume we are dealing with supervised binary classification problem with multiview data. A training data set of $n$ data points is given, $X = \{\theta_1, \cdots, \theta_n\}$. $\theta_i=(\bfx_i^1, \cdots, \bfx_i^m, y_i)$ is the $i$-th data point. The information of each data point is composed of feature vectors of $m$ views, and a binary class label $y_i$. $\bfx_i^j\in \mathbb{R}^{d_j}$ is the $d_j$-dimensional feature vector of the $j$-th view of the $i$-th data point, and $y_i\in \{+1,-1\}$ is a the binary class label of the $i$-th data point. The problem of supervised multiview learning is to learn a predictive model from the training set, which can predict a binary class label from the multiview input of a test data point. We assume there is an intact vector $\bfz_i\in \mathbb{R}^d$ for the $i$-th data point, and its $j$-th view $\bfx_i^j$ can be reconstructed by a linear transformation,

\begin{equation}
\begin{aligned}
\bfx_i^j \leftarrow W_j \bfz_i,
\end{aligned}
\end{equation}
where $W_j\in R^{d_j\times d}$ is the view-conditional linear transformation matrix for the $j$-th view. Please the view-conditional transformation matrix for the same view of all the data points is the same. By learning both the $W_j$ and $\bfz_i$, we can recover the hidden intact vector for the $i$-th data point, $\bfz_i$, and use it for classification problem. To this end, we propose to minimize the reconstruction error. The reconstruction error is measured by Cauchy error estimator, $E(\bfx_i^j, W_j \bfz_i)$,

\begin{equation}
\begin{aligned}
E(\bfx_i^j, W_j \bfz_i) = \log \left (
1 + \frac{\left \| \bfx_i^j - W_j \bfz_i \right \|_2^2}{c^2}
 \right ).
\end{aligned}
\end{equation}
This error estimator has been shown to be robust, and it also provides a offset. We propose to minimize this  error estimator over all data points and all views with regard to both $\bfz_i, i=1, \cdots, n$, and $W_j, j=1, \cdots, m$,

\begin{equation}
\label{equ:reconstruction}
\begin{aligned}
\min_{\bfz_i|_{i=1}^n,W_j|_{j=1}^m}
\left \{ \sum_{i=1}^n \sum_{j=1}^m E(\bfx_i^j, W_j \bfz_i)
=
\sum_{i=1}^n \sum_{j=1}^m \log \left (
1 + \frac{\left \| \bfx_i^j - W_j \bfz_i \right \|_2^2}{c^2}
 \right )
\right \}
\end{aligned}
\end{equation}

Moreover, we also assume that the intact feature vectors of the data points are discriminative, and presents the class information, thus the intact feature vectors can minimize a classification loss function of the data set. We propose to learn the intact feature vector of the $i$-th data point by jointly learning a liner classifier to predict its class label, $y_i$. The classifier is designed as linear function,

\begin{equation}
\begin{aligned}
y_i \leftarrow \bfomega^\top \bfz_i
\end{aligned}
\end{equation}
The usage of a linear function as the classifier is motive by the work of Fan and Tang \cite{fan2010enhanced}. Fan and Tang \cite{fan2010enhanced} proposed to use a linear classifier to maximize the area under the ROC Curve (AUC) for the problem of imbalance learning and cost sensitive learning. Fan and Tang \cite{fan2010enhanced} found that a linear classifier used to maximize AUC searches an optimal solution in a very constrained space, and enhance the maximum AUC linear classifier by extending its searching in the solution space, and improving the way to use the structure of the classifier. Thus the linear classifier has been proven to be effective in the optimizing AUC by Fan and Tang \cite{fan2010enhanced}, it inspires us to use it to learn an effective classifier in the intact vector space. The classification error can be measured by the hinge loss function,

\begin{equation}
\begin{aligned}
L(y_i, \bfomega^\top \bfz_i) = \max(0, 1 - y_i \bfomega^\top \bfz_i).
\end{aligned}
\end{equation}
This the optimization of this loss function can obtain a large margin classifier. To learn the optimal classifier and the discriminative intact feature vectors, we propose to minimize the classifier loss measured by the hinge loss function of the classification result over all the training data points,

\begin{equation}
\label{equ:classifier}
\begin{aligned}
\min_{\bfz_i|_{i=1}^n, \bfomega} \left \{ \sum_{i=1}^n
L(y_i, \bfomega^\top \bfz_i) = \sum_{i=1}^n  \max(0, 1 - y_i \bfomega^\top \bfz_i)
\right \}
\end{aligned}
\end{equation}

Moreover, to prevent the problem of over-fitting of variables, we propose to minimize the squared $\ell_2$ norm of the variables to regularize the learning $\bfz_i$, $W_j$, and $\bfomega$,

\begin{equation}
\label{equ:regular}
\begin{aligned}
\min_{\bfz_i|_{i=1}^n, W_j|_{j=1}^m,\bfomega} \left \{
R(\bfz_i|_{i=1}^n, W_j|_{j=1}^m,\bfomega) =  \sum_{i=1}^n \|\bfz_i\|_2^2 +  \sum_{j=1}^m \|W_j\|_2^2 +  \|\bfomega\|_2^2
\right \}.
\end{aligned}
\end{equation}

Our overall learning problem is obtained by considering both the problems of view-conditional reconstruction in (\ref{equ:reconstruction}), and classifier learning in the intact space in (\ref{equ:classifier}),

\begin{equation}
\label{equ:objective}
\begin{aligned}
\min_{\bfz_i|_{i=1}^n,W_j|_{j=1}^m,\bfomega}
&
\left \{
\sum_{i=1}^n \sum_{j=1}^m E(\bfx_i^j, W_j \bfz_i) + \alpha L(y_i, \bfomega^\top \bfz_i) + \gamma R(\bfz_i|_{i=1}^n, W_j|_{j=1}^m,\bfomega)
\right .
\\
&\left .
=
\sum_{i=1}^n \sum_{j=1}^m \log \left (
1 + \frac{\left \| \bfx_i^j - W_j \bfz_i \right \|_2^2}{c^2}
 \right )
 \right .\\
& + \alpha
\sum_{i=1}^n  \max(0, 1 - y_i \bfomega^\top \bfz_i)\\
&
\left .
+ \gamma \left (
 \sum_{i=1}^n \|\bfz_i\|_2^2 +  \sum_{j=1}^m \|W_j\|_2^2 +  \|\bfomega\|_2^2
\right )
\vphantom{\sum_{i=1}^n \sum_{j=1}^m E(\bfx_i^j, W_j \bfz_i) }
\right \},
\end{aligned}
\end{equation}
where $\alpha$ is a tradeoff parameter to balance the view-conditional reconstruction terms and the classification error terms, and $\gamma$ is a tradeoff parameter to balance the view-conditional reconstruction terms and the regularization terms. By optimizing this problem, we can learn intact feature vectors which can present the multiview inputs of the data points, and also is discriminative.

\subsection{Optimization}

To solve the optimization problem in (\ref{equ:objective}), we propose to use the alternate optimization strategy. The optimization is conducted in an iterative algorithm. When one variable is considered, the others are fixed. After one variable is updated, it will be fixed in the next iteration when other variable is updated. In the following subsections, we will discuss how to update each variable.

\subsubsection{Updating $\bfz_i$}

When we want to update $\bfz_i$, we only consider this single variable, while fix all other variables. Thus we have the following optimization problem,

\begin{equation}
\label{equ:zi}
\begin{aligned}
\min_{\bfz_i}
&
\left \{
\sum_{j=1}^m \log \left (
1 + \frac{\left \| \bfx_i^j - W_j \bfz_i \right \|_2^2}{c^2}
 \right )
 +
\alpha
\max(0, 1 - y_i \bfomega^\top \bfz_i) + \gamma  \|\bfz_i\|_2^2
\right \}.
\end{aligned}
\end{equation}
The second term $\max(0, 1 - y_i \bfomega^\top \bfz_i)$ is not a convex function, and it is hard to optimize it directly. Thus we rewrite it as follows,

\begin{equation}
\label{equ:hinge}
\begin{aligned}
\max(0, 1 - y_i \bfomega^\top \bfz_i)
=
\left\{\begin{matrix}
1 - y_i \bfomega^\top \bfz_i, &if~ 1 - y_i \bfomega^\top \bfz_i > 0\\
0, &otherwise.
\end{matrix}\right.
\end{aligned}
\end{equation}
We define a indicator variable, $\beta_i$, to indicate which of the above cases is true,

\begin{equation}
\label{equ:beta}
\begin{aligned}
\beta_i
=
\left\{\begin{matrix}
1, &if~ 1 - y_i \bfomega^\top \bfz_i > 0\\
0, &otherwise,
\end{matrix}\right.
\end{aligned}
\end{equation}
and rewrite (\ref{equ:hinge}) as follows,

\begin{equation}
\label{equ:hinge1}
\begin{aligned}
\max(0, 1 - y_i \bfomega^\top \bfz_i)
=
\beta_i  \left (
1 - y_i \bfomega^\top \bfz_i
\right )
\end{aligned}
\end{equation}
Please note that $\beta_i$ is also a function of $\bfz_i$, however, we first update it by using $\bfz_i$ solved in previous iteration, and then fix it to update $\bfz_i$ in current iteration. In this way, (\ref{equ:zi}) is rewritten as

\begin{equation}
\label{equ:zi1}
\begin{aligned}
\min_{\bfz_i}
&
\left \{
\sum_{j=1}^m \log \left (
1 + \frac{\left \| \bfx_i^j - W_j \bfz_i \right \|_2^2}{c^2}
 \right )
 +
\alpha
\beta_i  \left (
1 - y_i \bfomega^\top \bfz_i
\right ) + \gamma  \|\bfz_i\|_2^2 = g(\bfz_i)
\right \},
\end{aligned}
\end{equation}
where $g(\bfz_i)$ is the objective of this optimization problem. To seek the minimization of $g(\bfz_i)$, we use gradient descent algorithm. This algorithm update $\bfz_i$ by descending it to the direction of gradient of $g(\bfz_i)$,

\begin{equation}
\label{equ:zi1_gradient}
\begin{aligned}
\bfz_i \leftarrow \bfz_i - \mu \nabla g(\bfz_i),
\end{aligned}
\end{equation}
where $\mu$ is the descent step, and $\nabla g(\bfz_i)$ is the gradient function of $g(\bfz_i)$. We set $\nabla g(\bfz_i)$ as the partial derivative of $g(\bfz_i)$ with regard to $\bfz_i$,

\begin{equation}
\label{equ:derivative}
\begin{aligned}
\nabla g(\bfz_i)
&=
\frac{\partial g(\bfz_i)}{\partial \bfz_i}
=
\sum_{j=1}^m \frac{\frac{2 W_j^\top( \bfx_i^j - W_j \bfz_i  )}{c^2}}{\left (
1 + \frac{\left \| \bfx_i^j - W_j \bfz_i \right \|_2^2}{c^2} \right )
}
 -
\alpha
\beta_i  y_i \bfomega  +  \gamma \bfz_i\\
&=
\sum_{j=1}^m \frac{2 W_j^\top( \bfx_i^j - W_j \bfz_i  )}{\left (
c^2 + \left \| \bfx_i^j - W_j \bfz_i \right \|_2^2\right )}
 -
\alpha
\beta_i  y_i \bfomega +  \gamma \bfz_i.
\end{aligned}
\end{equation}
By substituting (\ref{equ:derivative}) to (\ref{equ:zi1_gradient}), we have the final updating rule of $\bfz_i$,

\begin{equation}
\label{equ:zi1_gradient1}
\begin{aligned}
\bfz_i \leftarrow \bfz_i - \mu
\left (
\sum_{j=1}^m \frac{2 W_j^\top( \bfx_i^j - W_j \bfz_i  )}{\left (
c^2 + \left \| \bfx_i^j - W_j \bfz_i \right \|_2^2\right )}
 -
\alpha
\beta_i  y_i \bfomega +  \gamma \bfz_i
\right ).
\end{aligned}
\end{equation}

\subsubsection{Updating $W_j$}

When we want to optimize $W_j$, we fix all other variables and only consider $W_j$ itself. The optimization problem is changed to the follows,

\begin{equation}
\label{equ:Wj}
\begin{aligned}
\min_{W_j}
&
\left \{
\sum_{i=1}^{n} \log \left ( 1 + \frac{\left \| \bfx_i^j - W_j \bfz_i \right \|_2^2}{c^2}
 \right ) + \gamma \|W_j\|_2^2= f(W_j) \right \}.
\end{aligned}
\end{equation}
where $f(W_j)$ is the objective function of this problem. To solve this problem, we also update $W_j$ by using the gradient descent algorithm,

\begin{equation}
\label{equ:Wj_gradient}
\begin{aligned}
W_j \leftarrow W_j - \mu \nabla f(W_j),
\end{aligned}
\end{equation}
where $\nabla f(W_j)$ is the gradient function of $f(W_j)$,

\begin{equation}
\label{equ:f_gradient}
\begin{aligned}
\nabla f(W_j)
&= \frac{\partial f(W_j)}{\partial W_j} =
\sum_{i=1}^{n} \frac{\frac{2(\bfx_i^j - W_j\bfz_i)\bfz_i^\top}{c^2}}{ \left ( 1 + \frac{\left \| \bfx_i^j - W_j \bfz_i \right \|_2^2}{c^2}
 \right )}+ \gamma W_j\\
& = \sum_{i=1}^{n} \frac{2(\bfx_i^j - W_j\bfz_i)\bfz_i^\top}{ \left ( c^2 + \left \| \bfx_i^j - W_j \bfz_i \right \|_2^2  \right )} + \gamma W_j.
\end{aligned}
\end{equation}
Substituting (\ref{equ:f_gradient}) to (\ref{equ:Wj_gradient}), we have the final updating rule of $W_j$,

\begin{equation}
\label{equ:Wj_gradient1}
\begin{aligned}
W_j \leftarrow W_j - \mu \left (
\sum_{i=1}^{n} \frac{2(\bfx_i^j - W_j\bfz_i)\bfz_i^\top}{ \left ( c^2 + \left \| \bfx_i^j - W_j \bfz_i \right \|_2^2  \right )}+ \gamma W_j
\right ).
\end{aligned}
\end{equation}

\subsubsection{Updating $\bfomega$}

When we want to update $\bfomega$ to minimize the objective function of (\ref{equ:objective}), we fix the other variables, and only consider $\bfomega$. Thus the problem in (\ref{equ:objective}) is transferred to

\begin{equation}
\label{equ:objective}
\begin{aligned}
\min_{\bfomega}
&
\left \{ \alpha\sum_{i=1}^n
\beta_i  \left (
1 - y_i \bfomega^\top \bfz_i
\right ) + \gamma \|\bfomega\|_2^2 = h(\bfomega)
\right \}.
\end{aligned}
\end{equation}
Please note that $\beta_i$ is actually a function of $\bfomega$. However, similar the strategy to solve $\bfz_i$, we also update it according to $\bfomega$ solved in previous iteration, and fix it to update $\bfomega$ in current iteration. When $\beta_i, i=1, \cdots, n$  are fixed, we update $\bfomega$ to minimize $h(\bfomega)$ by using the gradient descent algorithm,

\begin{equation}
\label{equ:beta_gradient}
\begin{aligned}
\bfomega \leftarrow \bfomega - \mu \nabla h(\bfomega),
\end{aligned}
\end{equation}
where $\nabla h(\bfomega)$ is the gradient function of $h(\bfomega)$, and it is defined as follows,

\begin{equation}
\label{equ:h_gradient}
\begin{aligned}
\nabla h(\bfomega) = \frac{\partial h(\bfomega)}{\partial \bfomega}
=- \alpha \sum_{i=1}^n \beta_i y_i \bfz_i + \gamma \bfomega.
\end{aligned}
\end{equation}
By substituting it to (\ref{equ:beta_gradient}), we have the final updating rule for $\bfomega$,

\begin{equation}
\label{equ:beta_gradient}
\begin{aligned}
\bfomega \leftarrow \bfomega - \mu \left (
- \alpha \sum_{i=1}^n \beta_i y_i \bfz_i + \gamma \bfomega
\right ).
\end{aligned}
\end{equation}

\subsection{Iterative algorithm}

After we have the updating rules of all the variables, we can design an iterative algorithm for the learning problem. This iterative algorithm has one outer FOR loop, and two inner FOR loops. The outer FOR loop is corresponding to the main iterations. The two inner FOR loops are corresponding to the updating of $n$ intact feature vectors of $n$ data points, and the updating of $m$ view-conditional transformation matrices. The algorithm is given in Algorithm 1. {The iteration number $T$ is determined by cross-validation in our experiments. }

\begin{itemize}
\item \textbf{Algorithm 1}. Iterative algorithm for multiview intact and single-view classifier learning (MISC).
\item \textbf{Input}: Training data set, $(\bfx_1^1, \cdots, \bfx_1^m, y_1), \cdots, (\bfx_n^1, \cdots, \bfx_n^m, y_n)$.
\item \textbf{Input}: Tradeoff parameters, $\alpha$ and $\gamma$.
\item \textbf{Input}: Maximum iteration number, $T$.

\item \textbf{Initialization}: $\bfz_i^0, i=1,\cdots,n$, $W_j^0,j=1,\cdots, m$ and $\bfomega^0$.

\item \textbf{For $t=1,\cdots, T$}

\begin{itemize}
\item Update descent step, $\mu^t \leftarrow \frac{1}{t}$

\item \textbf{For $i=1,\cdots,n$}

Update $\beta_i^t$ as follows,

\begin{equation}
\begin{aligned}
\beta_i^t
=
\left\{\begin{matrix}
1, &if~ 1 - y_i {\bfomega^{t-1}}^\top \bfz_i^{t-1} > 0\\
0, &otherwise.
\end{matrix}\right.
\end{aligned}
\end{equation}

Update $\bfz_i^t$ by fixing $W_j^{t-1}, j=1,\cdots, m$, $\beta_i^{t-1}$ and $\bfomega^{t-1}$,

\begin{equation}
\begin{aligned}
\bfz_i^t \leftarrow \bfz_i^{t-1} - \mu^t
\left (
\sum_{j=1}^m \frac{2 {W_j^{t-1}}^\top( \bfx_i^j - {W_j^{t-1}} \bfz_i^{t-1}  )}{\left (
c^2 + \left \| \bfx_i^j - {W_j^{t-1}} \bfz_i^{t-1} \right \|_2^2\right )}
 -
\alpha
\beta_i^t  y_i \bfomega^{t-1} +  \gamma \bfz_i^{t-1}
\right ).
\end{aligned}
\end{equation}

\item \textbf{End of For}

\item \textbf{For $j=1,\cdots,m$}

Update $W_j^t$ by fixing $\bfz_i^{t}, i=1,\cdots, m$,

\begin{equation}
\begin{aligned}
W_j^t \leftarrow W_j^{t-1} - \mu^t \left (
\sum_{i=1}^{n} \frac{2(\bfx_i^j - {W_j^{t-1}}\bfz_i^{t}){\bfz_i^t}^\top}{ \left ( c^2 + \left \| \bfx_i^j - W_j^{t-1} \bfz_i^t \right \|_2^2  \right )}+ \gamma W_j^{t-1}
\right ).
\end{aligned}
\end{equation}

\item \textbf{End of For}

\item Update $\bfomega^t$ by fixing $\beta_i^t, i=1, \cdots, n$ and $\bfz_i^t, i=1, \cdots, n$,

\begin{equation}
\begin{aligned}
\bfomega^t \leftarrow \bfomega^{t-1} - \mu^t \left (
- \alpha \sum_{i=1}^n \beta_i^t y_i \bfz_i + \gamma \bfomega^{t-1}
\right ).
\end{aligned}
\end{equation}

\end{itemize}

\item \textbf{End of For}

\item \textbf{Output}: $W_j^T, j=1, \cdots, m$, $\bfz_i^T, i=1, \cdots, n$, and $\bfomega^T$.
\end{itemize}

As we can see from the algorithm, in the main FOR loop, descent step variable, $\mu$, is firstly updated, and then the hinge loss indicator variables, $\beta_i, i=1,\cdots,n$ and the intact feature vectors are updated. The view-conditional transformation matrices, $W_j, j=1, \cdots, m$ are updated, and finally, the classifier parameter $\bfomega$ are updated.

\newpage

\section{Experiments}
\label{sec:experiments}

In this section, we will evaluate the proposed algorithm on a few real-world supervised multiview learning problems experimentally.

\subsection{Benchmark data sets}

\subsubsection{PASCAL VOC 07 data set}

The first data set used in the experiment is the PASCAL VOC 07 data set \cite{Jaszewski2015}. In this data set, there are 9,963 images of 20 different object classes. Each image is presented by two different view, which are visual view, and tag view. To extract the feature vector from the visual view of an image, we extract local visual features, SIFT, from the image, and represent the local features as a histogram. To extract the feature vector from the tag view from the image, we use the histogram vector of user tags of the image as the feature vector.

\subsubsection{CiteSeer data set}

The second data set is the CiteSeer data set \cite{Williams201468}. In this data set, there are 3,312 documents of 6 classes. Each document has three views, which are the text view, inbound reference view, and outbound reference view.

\subsubsection{HMDB data set}

The third data set is the HMDB dataset, which is a video database of human motion recognition problem \cite{Kuehne20112556}. In this data set, there are 6,849 video clips of 51 action classes. To present each video clip, we extract 3D Harris corners, and present them by two different types of local features, which are the histogram of oriented gradient (HOG) and histogram of oriented flow (HOF).  We further represent each clip by two feature vectors of two views, which are the histograms of HOG and HOF.

\subsection{Experiment protocols}

To conduct the experiments, we split each data set into 10 non-overlapping folds, and use the 10-fold cross-validation to perform the training-testing procedure. Each fold is used as a test set in turn, and the remaining 9 folds are used as the training sets. The proposed algorithm is performed to the training set to obtain the view-conditional transformation matrices, and the classifier parameter. Then the learned view-conditional transformation matrices and the classifier parameter are used to represent and classify the data points in the test set. To handle the multiple class problem, we use the one-vs-all strategy.

\subsection{Performance measures}

To measure the classification performance over the test set, we use the classification accuracy. The classification accuracy is defined as follows,

\begin{equation}
\begin{aligned}
Classification~accuracy=
\frac{Number~of~correctly~classified~test~data~points}{Number~of~total~test~data~points}.
\end{aligned}
\end{equation}
It is obvious that a better algorithm should be able to obtain a higher  classification accuracy.

\subsection{Experiment results}

In this experiment, we first study the sensitivity of the algorithm to the parameters, which are $\alpha$ and $\gamma$.

\subsubsection{Sensitivity to parameters}

\begin{figure}[h!]
  \centering
  \includegraphics[width=\textwidth]{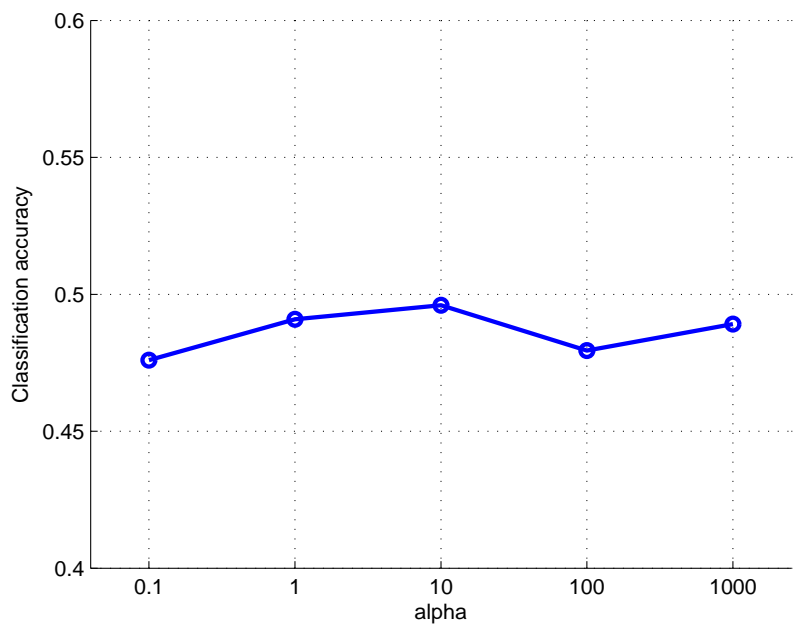}\\
  \caption{Sensitivity curve of $\alpha$ over PASCAL VOC 07 data set.}
  \label{Fig_alpha150805}
\end{figure}

\begin{figure}[h!]
  \centering
  \includegraphics[width=\textwidth]{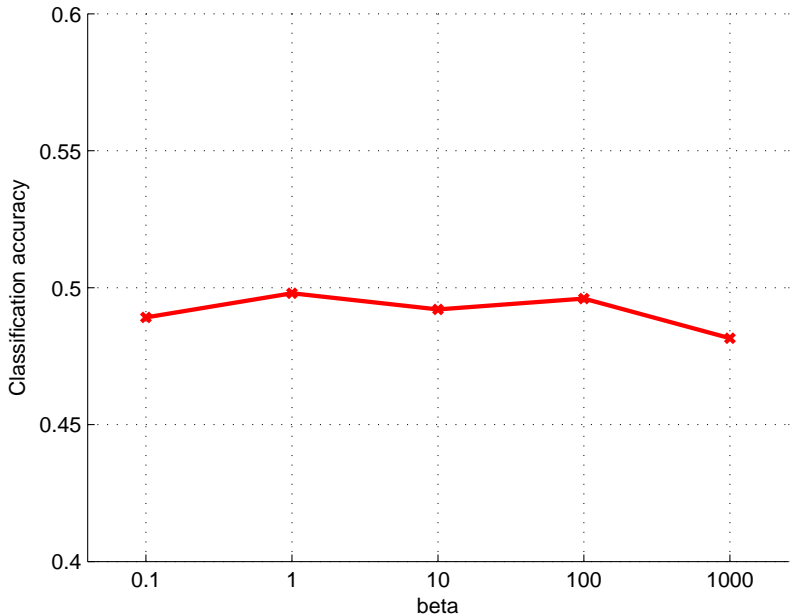}\\
  \caption{Sensitivity curve of $\beta$ over PASCAL VOC 07 data set.}
  \label{Fig_beta150807}
\end{figure}

To study the performance of the proposed algorithm with different tradeoff parameters, $\alpha$ and $\beta$. We perform the algorithm by using the parameters of values $0.1, 1, 10, 100$ and $1000$, and measured the performance of different parameters. Fig. \ref{Fig_alpha150805} illustrates the performance on the PASCAL VOC 07 data set with respect to different tradeoff parameter $\alpha$. The proposed algorithm achieves a stable performance in all the settings of parameter $\alpha$. In Fig. \ref{Fig_beta150807}, the performance against different tradeoff parameter $\beta$ is also shown. From this figure, we can also see that the algorithm is stable tot he changes of value of $\beta$. This suggests that MISC is not sensitive to the changes of tradeoff parameters.

\subsubsection{Comparison to state-of-the-art algorithms}

\begin{figure}[h!]
\centering
\includegraphics[width=\textwidth]{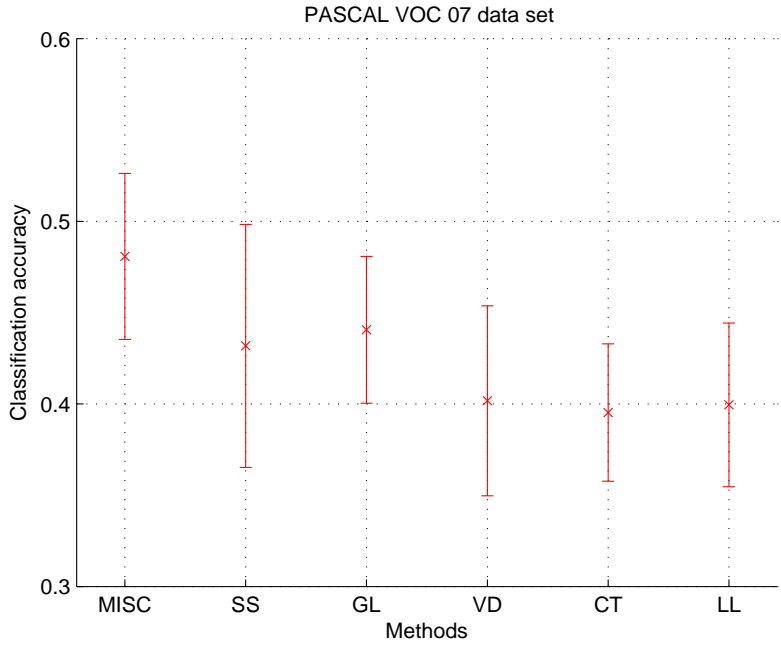}\\
\caption{Results of comparison of different algorithms over PASCAL VOC'07 data set.}
\label{FigCompr1}
\end{figure}

\begin{figure}[h!]
\centering
\includegraphics[width=\textwidth]{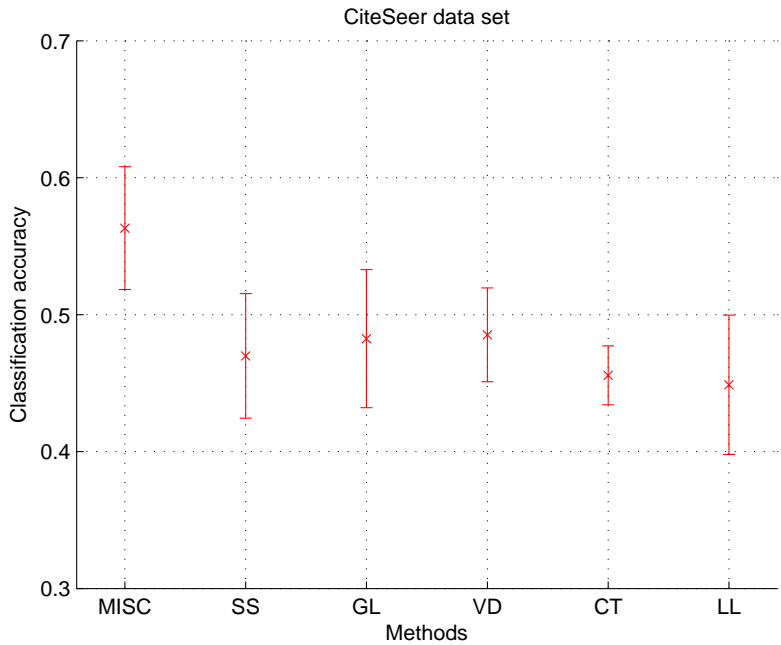}\\
\caption{Results of comparison of different algorithms over CiteSeer data set.}
\label{FigCompr2}
\end{figure}

\begin{figure}[h!]
\centering
\includegraphics[width=\textwidth]{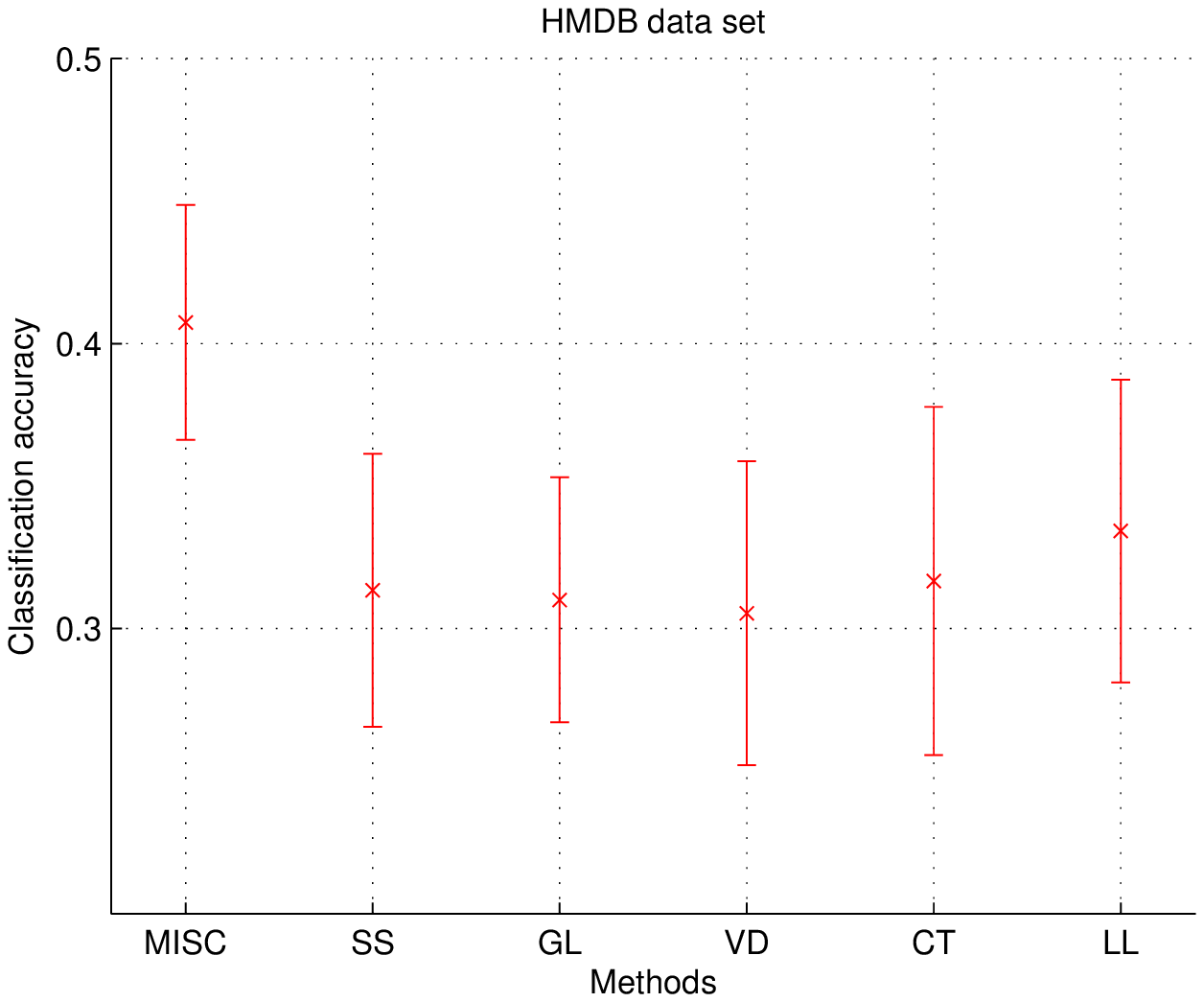}\\
\caption{Results of comparison of different algorithms over HMDB data set.}
\label{FigCompr3}
\end{figure}

We compare the proposed algorithm to the following methods, multiview learning algorithm using local learning (LL) proposed by Zhang et al. \cite{Zhang2008752}, multiview learning algorithm using co-training (CT) proposed by Sindhwani et al. \cite{sindhwani2005co}, multiview learning algorithm based on view disagreement (VD) proposed by Quadrianto \cite{Quadrianto2011425}, multiview learning algorithm with global consistency and local smoothness (GL) proposed by Zhai \cite{Zhai2012}, and multiview representation method using statistical subspace learning (SS) proposed by Chen et al. \cite{Chen20122365}. The error bars of the classification accuracy of the compared methods over three different data sets are given in Fig. \ref{FigCompr1}, Fig. \ref{FigCompr1} and Fig. \ref{FigCompr3}. From the figures, we find that the proposed method, MISC, stably outperforms other algorithms at all the data sets. Even on the most difficult data set, HMDB, the proposed method, MISC, also achieves an accuracy as high as about 0.4. The multiple view data are optimally combined by MISC to find the latent intact space and the optimal classifier in the corresponding intact space. The main reason for this is the robust property of the proposed algorithm. This algorithm has the ability to appropriately handle the complementary between multiple views, and learn a discriminative hidden intact space with help of classifier learning.

\section{Conclusions and future works}
\label{sec:conclusion}

We propose a novel multiview learning algorithm by learning intact vectors of the training data points and a classifier in the intact space. The intact vectors is assumed to be a hidden but critical vector for each data point, and we can obtain its multiple view feature vectors by view-conditional transformations. Moreover, we also assume that the intact vectors are discriminative, i.e., can be separated by a linear function according to their classes. We propose a novel optimization problem to model both the learning of intact vectors and classifier. An iterative algorithm is developed to solve this problem. This algorithm outperforms other multiview learning algorithms on benchmark data sets, and it also shows its stability over tradeoff parameters. In the future, we will study the potential to use the proposed algorithm for imbalanced data set with multi-view features \cite{fan2011margin,fan2010enhanced,chawla2004editorial}, and the usage of Bayesian network classifier instead of linear classifier to learn the intact vectors of multi-view data \cite{fan2014tightening,fan2014finding,fan2015improved}. Moreover, we will also investigate to use the proposed algorithm to solve the problems of bioinformatics \cite{wang2014computational,zhou2014biomarker,liu2013structure,peng2015modeling}, computer vision \cite{wang2015representing,wang2015image}, and multi-media data processing \cite{wang2015supervised,wang2015multiple}.

\section*{Acknowledgements}

This project was supported by the National Natural Science Foundation of China (Grant No. 61472172), and a research funding of Ludong University (Grant No. 27870301).


\end{document}